\documentclass{article} % For LaTeX2e

\usepackage[preprint, nonatbib]{neurips_2024}
\usepackage{booktabs,multirow,graphicx,amsthm,array}
\usepackage{wrapfig}
\usepackage{amsmath,amsfonts,bm}

\def\1{\bm{1}}

\DeclareMathAlphabet{\mathsfit}{\encodingdefault}{\sfdefault}{m}{sl}
\SetMathAlphabet{\mathsfit}{bold}{\encodingdefault}{\sfdefault}{bx}{n}

\usepackage{xcolor}
\usepackage{amsthm}
\usepackage{amsmath}

\usepackage{hyperref}
\definecolor{cvprblue}{rgb}{0.21,0.49,0.74}
\hypersetup{
    colorlinks=true, % Set to 'true' to enable colored links
    linkcolor=red,   % Color of internal links
    filecolor=magenta, % Color of file links
    urlcolor=cyan,    % Color of external urls
    citecolor=cvprblue,   % Color of citations
}

\usepackage{url}
\usepackage{color,xcolor,colortbl,wrapfig}  
\usepackage{pifont}
\usepackage{graphicx}
\usepackage{subcaption}

\RequirePackage{xspace}
\makeatletter
\DeclareRobustCommand\onedot{\futurelet\@let@token\@onedot}
\def\@onedot{\ifx\@let@token.\else.\null\fi\xspace}

\def\eg{\emph{e.g}\onedot}

\providecommand{\thetitle}{}
\let\oldtitle\title
\renewcommand{\title}[1]{\oldtitle{#1}\renewcommand{\thetitle}{#1}}
\newcommand{\maketitlesupplementary}{
    \newpage
    \begin{center}
        \Large
        \textbf{\thetitle}\\[0.5em] % Adjust the vertical space as needed
        Supplementary Material\\[1.0em]
    \end{center}
}

\newcommand{\bbx}{\mathbf{x}}

\title{Gamba: Marry Gaussian Splatting with Mamba for Single-View 3D Reconstruction}

\def\mystrut{\rule{0pt}{1.0\normalbaselineskip}}
\author{
\begin{tabular}{@{}c}
Qiuhong Shen$^1${\footnotemark[1]} \quad 
Zike Wu$^3${\footnotemark[1]} \quad 
Xuanyu Yi$^3${\footnotemark[1]}\quad  
Pan Zhou$^{2,4}${\footnotemark[2]}\mystrut  \quad 
Hanwang Zhang$^{3,5}$\quad \\ Shuicheng Yan$^5$\quad Xinchao Wang$^1${\footnotemark[2]}\mystrut \\
\end{tabular}\\
$^1$National University of Singapore\mystrut\quad
$^2$Singapore Management University\\
$^3$Nanyang Technological University \quad
$^4$Sea AI Lab\quad
$^5$Skywork AI\\
}

\begin{document}

\maketitle

\renewcommand{\thefootnote}{\fnsymbol{footnote}}
\footnotetext[1]{Equal Contribution}
\footnotetext[2]{Corresponding author :  Xinchao Wang (xinchao@nus.edu.sg) and Pan Zhou (panzhou@smu.edu.sg)}
\footnotetext[3]{Work partially done in Sea AI Lab and 2050 Research, Skywork AI}

\begin{abstract}

We tackle the challenge of efficiently reconstructing a 3D asset from a single image at millisecond speed. Existing methods for single-image 3D reconstruction are primarily based on Score Distillation Sampling (SDS) with Neural 3D representations. Despite promising results, these approaches encounter practical limitations due to lengthy optimizations and significant memory consumption. In this work, we introduce Gamba, an end-to-end 3D reconstruction model from a single-view image, emphasizing two main insights: (1) Efficient Backbone Design: introducing a Mamba-based GambaFormer network to model 3D Gaussian Splatting (3DGS) reconstruction as sequential prediction with linear scalability of token length, thereby accommodating a substantial number of Gaussians; (2) Robust Gaussian Constraints: deriving radial mask constraints from multi-view masks to eliminate the need for warmup supervision of 3D point clouds in training. We trained Gamba on Objaverse and assessed it against existing optimization-based and feed-forward 3D reconstruction approaches on the GSO Dataset, among which Gamba is the only end-to-end trained single-view reconstruction model with 3DGS. Experimental results demonstrate its competitive generation capabilities both qualitatively and quantitatively and highlight its remarkable speed: Gamba completes reconstruction within 0.05 seconds on a single NVIDIA A100 GPU, which is about $1,000\times$ faster than optimization-based methods.  Please see our project page at \url{https://florinshen.github.io/gamba-project}.
% We will make all code and results publicly available.

\end{abstract}

\section{Introduction}
% 3dgs作为一种prevalent的新型表征，有很大优势和潜力
%基于3dgs表征的生成目前大多数方法都是per-instance的optimization，如何feed forward remain problem  ---引出task 3dgs feedforward model  （rare explored) 
% main challenge是什么？ (sequence length, unordered token, optimization constrains, )
%我们提出了一种first end-end（very important）的解决方案--mamba 1) parameter constrain 2) mamba structure --densitifcation 序列化
% 
%实验+总结

% why mamba: 3dgs这个表征的reconstruction过程包含clone, split，是condition on之前的点的，mamba的序列化特性很适合

We tackle the challenge of efficiently reconstructing a 3D asset from a single image, an endeavor with substantial implications across diverse industrial sectors. This endeavor facilitates AR/VR content generation from a single snapshot and aids in the development of autonomous vehicle path planning through monocular perception~\cite{sun2023trosd,gul2019comprehensive,yi2023invariant}.

% \textcolor{blue}{[Weakness of Optimization-based and nerf-based methods]} 
%费时和费资源，可以参考LRM，LGM写法
Previous approaches to single-view 3D reconstruction have mainly been achieved through Score Distillation Sampling~(SDS)~\cite{poole2022dreamfusion}, which leverages pre-trained 2D diffusion models~\cite{graikos2022diffusion,rombach2022high} to guide optimization of the underlying representations of 3D assets. These optimization-based approaches have achieved remarkable success, known for their high-fidelity and generalizability. 
However, they require a time-consuming per-instance optimization process~\cite{stable-dreamfusion,wang2023prolificdreamer,wu2024consistent3d} to generate a single object and also suffer from artifacts such as the “multi-face” problem arising from bias in pre-trained 2D diffusion models~\cite{hong2023debiasing}. On the other hand, previous approaches predominantly utilized neural radiance fields (NeRF)~\cite{mildenhall2021nerf,barron2021mip}, which are equipped with high-dimensional multi-layer perception (MLP) and inefficient volume rendering~\cite{mildenhall2021nerf}. This computational complexity significantly limits practical applications on limited compute budgets. For instance, the Large reconstruction Model (LRM)~\cite{adobelrm} is confined to a resolution of 32 using a triplane-NeRF~\cite{Shue_2023_CVPR} representation, and the resolution of renderings is limited to 128 due to the bottleneck of online volume rendering.

\begin{figure}[!t]
\centering
\includegraphics[width=\linewidth]{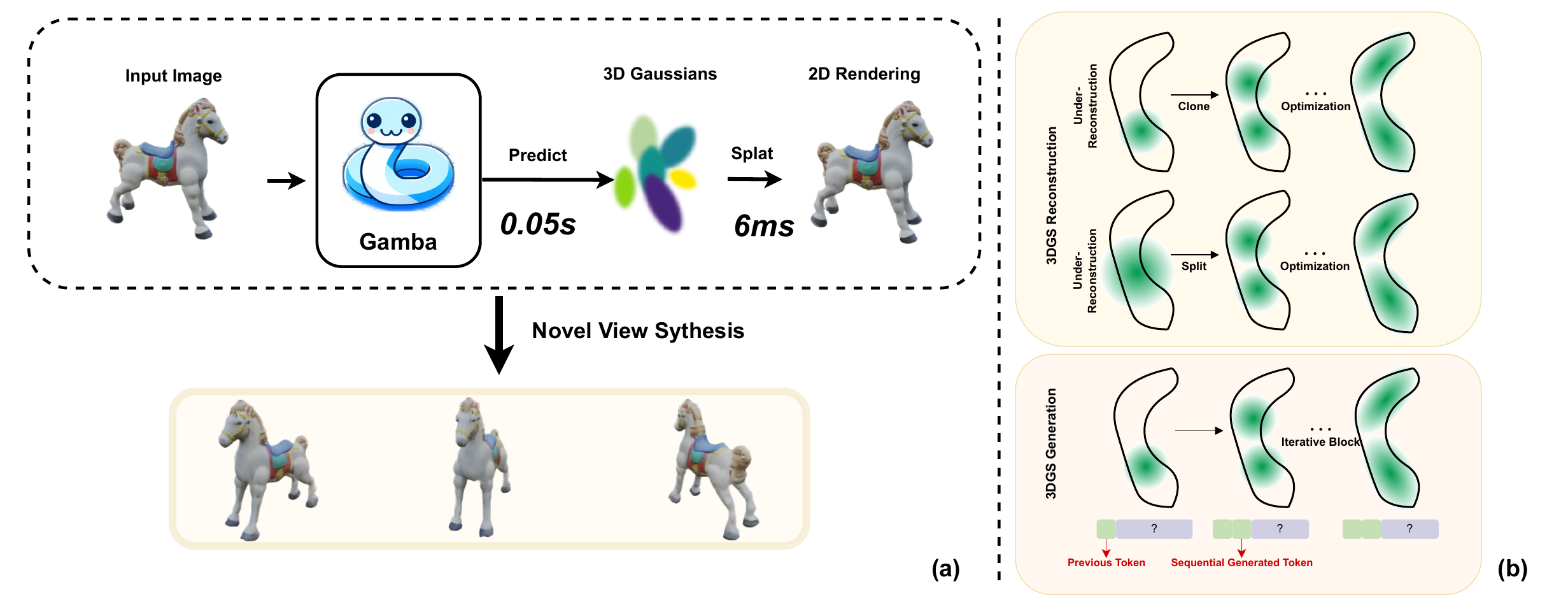}
\caption{(a): We propose Gamba, an end-to-end, feed-forward single-view reconstruction pipeline, which marries 3D Gaussian Splatting with Mamba to achieve fast reconstruction. (b): The relationship between the 3DGS iterative reconstruction and the Gamba sequential prediction pattern.}
\label{fig:1}
\vspace{-6mm}
\end{figure}

To address these challenges and thus achieve \textit{efficient} single-view 3D reconstruction, we are seeking an amortized reconstruction framework with the groundbreaking 3D Gaussian Splatting, notable for its memory-efficient and high-fidelity tiled rendering~\cite{kerbl20233d,zwicker2002ewa,chen2024survey,wang2024view}.
Despite recent exciting progress~\cite{tang2023dreamgaussian}, how to properly and immediately generate 3D Gaussians remains a less studied topic.
Recent prevalent 3D amortized generative models~\cite{adobelrm,wang2023pf,agg,dmv3d,zou2023triplane,li2023instant3d} predominantly use transformer-based architecture as their backbones~\cite{vaswani2017attention,peebles2023scalable}, but we argue that these widely used architectures are sub-optimal for generating 3DGS. 
%
% The crucial challenge stems from the fact that 3DGS requires a sufficient number of 3D Gaussians to accurately represent a single 3D object. 
% However, the computational complexity of attention mechanism in transformers increases quadratically with the number of tokens~\cite{vaswani2017attention}. Existing works~\cite{agg, zou2023triplane} struggles to build amortized 3DGS reconstruction model with transformer. They resort to a multi-stage training pagradigm to relax this bottleneck, but the performance is limited due to the insufficient Gaussians in the first stage. Furthermore, 3DGS has explicit, non-structural and discrete nature, making the simultaneous generation of 3DGS parameters a more challenging task compared to its NeRF counterparts.
%
The crucial challenge stems from the fact that 3DGS requires a sufficient number of 3D Gaussians to accurately represent a single 3D object. However, the computational complexity of the attention mechanism in transformers increases quadratically with the number of tokens~\cite{vaswani2017attention}. Existing works~\cite{agg, zou2023triplane} struggle to build amortized 3DGS reconstruction models with transformers. They resort to a multi-stage training paradigm to alleviate this bottleneck, but the performance is still limited due to the insufficient number of Gaussians in the first stage. Furthermore, 3DGS has explicit, non-structural, and discrete nature, making the simultaneous generation of 3DGS parameters a more challenging task compared to its Neural Radiance Fields (NeRF) counterparts.

% \textcolor{blue}{[Two contributions: 1. mamba arch 2. parameter constrain ]}
To tackle the above challenges, we start by revisiting the iterative 3DGS reconstruction  from posed multi-view 
images. The analysis presented in Figure~\ref{fig:1}(b) reveals that the densification during the
iterative 3DGS reconstruction can be conceptualized as a sequential prediction based on previously predicted tokens. With this insight, we introduce a novel architecture for \textit{end-to-end} 3DGS reconstruction dubbed Gaussian Mamba (Gamba), which is built upon a new scalable sequential network,
Mamba~\cite{gu2023mamba}. Our Gamba enables context-dependent reasoning and scales linearly with sequence (token) length, allowing it to efficiently mimic the inherent process of 3DGS reconstruction when reconstructing 3D assets enriched with a sufficient number of 3D Gaussians. 
% 
% Moreover, we train Gamba on the large-scale Objaverse dataset~\cite{deitke2023objaverse} with a robust training strategy in an end-to-end manner, and as its core, we explore radial mask constraint to supervise the position of Gaussians decently free of the need of explicit point cloud generative model. Due to its feed-forward, end-to-end architecture and efficient rendering, Gamba is exceptionally fast, requiring only about 1 seconds to generate a 3D asset and 6 ms for novel view synthesis, which is 5000× faster than previous optimization-based methods~\cite{wu2024consistent3d,weng2023consistent123,qian2023magic123} while achieving comparable generation quality.
% 
Moreover, we train Gamba on the large-scale Objaverse dataset~\cite{deitke2023objaverse} with a robust training constraints in an end-to-end manner. At its core, the model employs a radial mask constraint to supervise the placement of Gaussians effectively, thereby eliminating the need for an explicit point cloud supervision and multi-stage training in previous work~\cite{zou2023triplane, agg}. Due to its feed-forward, end-to-end architecture, combined with efficient rendering of 3DGS, Gamba achieves remarkable speed. It requires only about 0.05 seconds to generate a 3D asset and 6 ms for synthesizing novel views, which is $1000\times$ faster than previous optimization-based methods~\cite{weng2023consistent123,qian2023magic123} still delivering comparable quality in reconstruction outputs.

% Note that our model is only pretrained on multi-view image datasets, which is more scalable, and with more freedom to capture the intricate 3D parameter distributions.

We demonstrate the superiority of Gamba on the wide range of single images and decent evaluation on the Google Scanned Object (GSO) dataset~\cite{downs2022google}.
Both qualitative and quantitative experiments clearly indicate that Gamba can instantly generate high-quality and diverse 3D assets from a single image, continuously outperforming other state-of-the-art methods. In summary, we make three-fold contributions: 

\begin{itemize}
% \setlength{\itemsep}{2pt}
% \setlength{\parsep}{0pt}
% \setlength{\parskip}{0pt}
% \vspace{-1em}

\item We introduce GambaFormer, a simple Mamba-based reconstructor to process 3D Gaussian Splatting, which has global context length with linear complexity. 
\item Integrated with GambaFormer and robust 3DGS constraint, we present Gamba, an end-to-end 3DGS reconstruction pipeline for efficient single-view reconstruction.
\item Extensive experiments show that Gamba outperforms the state-of-the-art baselines in terms of reconstruction quality and speed.

\end{itemize}

\section{Related Works}

\noindent\textbf{Amortized 3D Generation.}
Amortized 3D generation is able to instantly generate 3D assets in a feed-forward manner after training on large-scale 3D datasets~\cite{wu2023omniobject3d,deitke2023objaverse,yu2023mvimgnet}, in contrast to tedious SDS-based optimization methods~\cite{wu2024consistent3d,lin2023magic3d,weng2023consistent123,threestudio2023,stable-dreamfusion}. Previous works~\cite{nichol2022point,nash2020polygen} married de-noising diffusion models with various 3D explicit representations (\eg, point cloud and mesh), which suffers from lack of generalizablity and low texture quality. Recently, pioneered by LRM~\cite{adobelrm}, several works utilize the capacity and scalability of the transformer~\cite{peebles2023scalable} and propose a full transformer-based regression model to decode a NeRF representation from triplane features. The following works extend LRM to predict multi-view images~\cite{li2023instant3d}, combine with diffusion~\cite{dmv3d}, and pose estimation~\cite{wang2023pf}. However, their triplane-NeRF representation is restricted to inefficient volume rendering and relatively low resolution with blurred textures. Gamba instead seeks to train an efficient feed-forward model marrying Gaussian splatting with Mamba for single-view 3D reconstruction.

\noindent\textbf{Gaussian Splatting for 3D Generation.}
The explicit nature of 3DGS facilitates real-time rendering capabilities and unprecedented levels of control and editability, making it highly relevant for 3D generation. Several works have effectively utilized 3DGS in conjunction with optimization-based 3D generation~\cite{wu2024consistent3d,poole2022dreamfusion,lin2023magic3d}. For example, DreamGaussian~\cite{tang2023dreamgaussian} utilizes 3D Gaussian as an efficient 3D representation that supports real-time high-resolution rendering via rasterization.
Despite the acceleration achieved, generating high-fidelity 3D Gaussians using such optimization-based methods still requires several minutes and a large computational memory demand. TriplaneGaussian~\cite{zou2023triplane} extends the LRM architecture with a hybrid triplane-Gaussian representation. AGG~\cite{agg} decomposes the geometry and texture generation task to
produce coarse 3D Gaussians, further improving its fidelity through Gaussian Super Resolution. Splatter image~\cite{szymanowicz2023splatter} and PixelSplat~\cite{charatan2023pixelsplat} propose to predict 3D Gaussians as pixels on the output feature map of two-view images. LGM~\cite{tang2024lgm} generates high-resolution 3D Gaussians by fusing information from multi-view images generated by existing multi-view diffusion models~\cite{shi2023mvdream,wang2023imagedream} with an asymmetric U-Net. Among them, our Gamba demonstrates its superiority and structural elegance with \textit{single image} as input and an \textit{end-to-end}, \textit{single-stage}, feed-forward manner.

\section{Method}
\begin{figure}[h]
\centering
\includegraphics[width=0.98\linewidth]{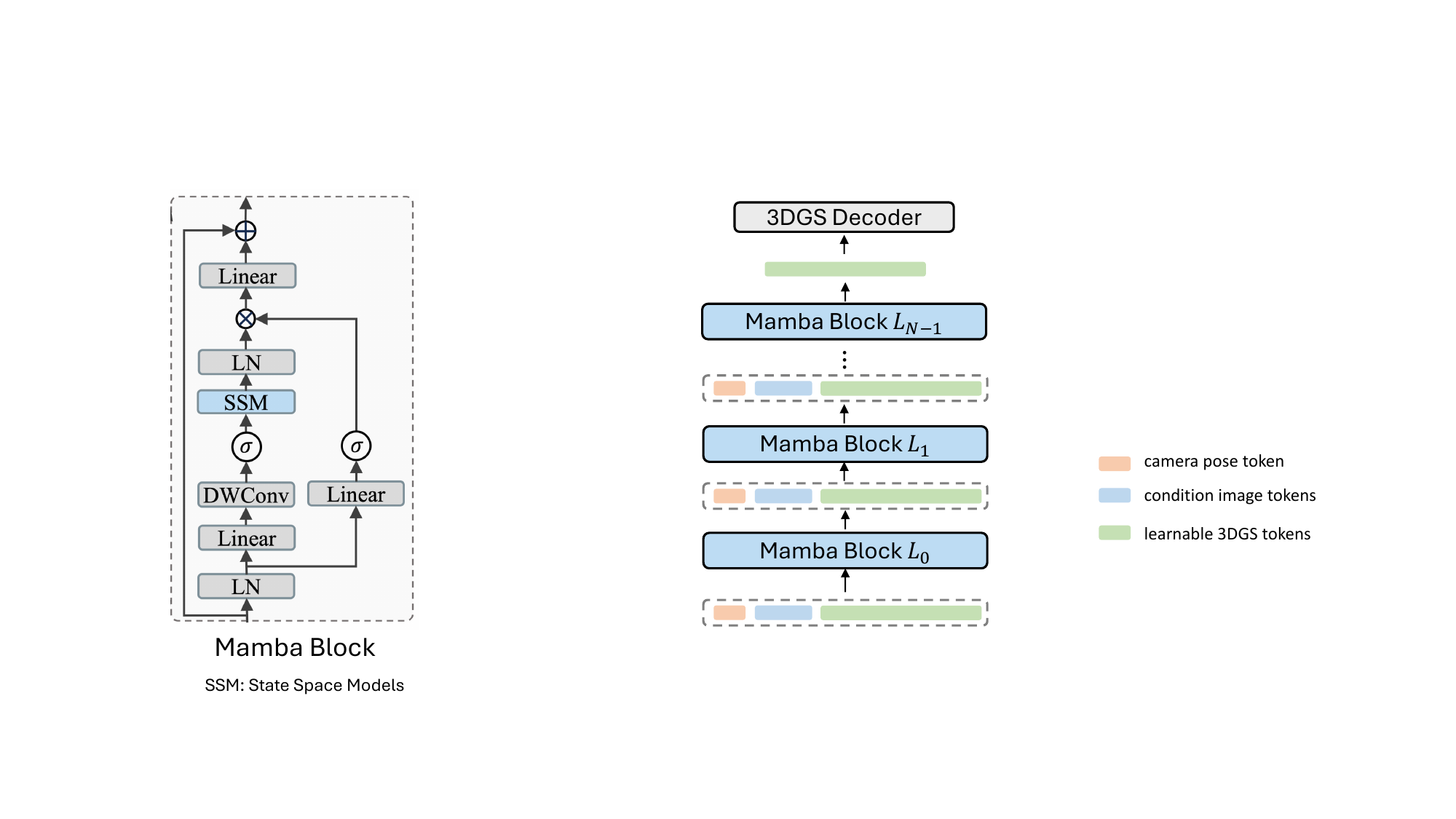}
\caption{Overall architecture of Gamba. Gamba takes a single view image and its camera pose as input to predict the 3D Gaussian Splatting of the given subject. Training supervision is only applied on the rendered multi-view images through reconstruction loss. }
\label{fig:overall_gambaformer}
\end{figure}

In this section, we detail our proposed single-view 3D reconstruction pipeline, which incorporates 3D Gaussian Splatting (3DGS) as depicted in Figure~\ref{fig:overall_gambaformer}(a), dubbed as ``Gamba." The core component of this pipeline is the GambaFormer, which predicts 3D Gaussians from a single image input (see Sec.~\ref{gambaformer}). We design elaborate constraints on the Gaussian parameters and a progressive training strategy, as discussed in Sec.~\ref{subsec:robust-train}, to achieve end-to-end training and high-fidelity reconstruction.

\subsection{Preliminary of 3D Gaussian Splatting}
3D Gaussian Splatting~(3DGS)~\cite{kerbl20233d} has gained prominence as an efficient explicit 3D representation, using anisotropic 3D Gaussians to achieve intricate modeling. Each Gaussian is defined by its 3D central position $\mu \in \mathbb{R}^3$, covariance matrix $\Sigma$, associated color $c \in \mathbb{R}^3$ (applicable when the degree of spherical harmonics is set to zero), and opacity $\alpha \in \mathbb{R}$. To be better optimized, the covariance matrix $\Sigma$ is constructed from a 3D scale $r\in \mathbb{R}^3$ and a rotation quaternion $q \in \mathbb{R}^4$. Generally, the $j$-th Gaussian can be collectively denoted as $\mathcal{G}_j = \{\mu_j, \alpha_j, r_j, q_j, c_j\}$. 3DGS projects Gaussians onto 2D images using a tile-based rasterization pipeline to support real-time rendering and differentiable optimization. This approach effectively controls the number of Gaussians through both adaptive densification and pruning of Gaussians.

% \subsection{Overview of Gamba Pipeline}
% Given a set of multi-view images and their paired camera pose $\{\bbx_i,\pi_i\}$ of a 3D object, Gamba first transforms the reference image $\bbx_{\text{ref}}$ and pose $\pi_{\text{ref}}$ into condition tokens. These tokens are then concatenated with the learnable 3DGS tokens to predict a set of 3D Gaussians. Subsequently, the predicted Gaussians are rendered into 2D multi-view images using the given camera poses $\{\pi\}$. This rendering process employs the differentiable rasterizer of 3DGS~\cite{kerbl20233d}, enabling direct supervision of the multi-view rendering output by the provided ground-truth images $\{\bbx\}$ at both reference and novel viewpoints through image space reconstruction loss.

 % Gamba first leverages a pre-trained visual transformer (DINO)~\cite{xx} to encode the reference image $I_{\text{ref}}$ to patch-wise feature tokens 

\subsection{GambaFormer} 
\label{gambaformer}
Given a set of multi-view images and their paired camera pose $\{\bbx_i,\pi_i\}$ of a 3D object, Gamba first transforms the reference image $\bbx_{\text{ref}}$ and pose $\pi_{\text{ref}}$ into condition tokens. These tokens are then concatenated with the learnable 3DGS tokens to predict a set of 3D Gaussians. Subsequently, the predicted Gaussians are rendered into 2D multi-view images using the given camera poses $\{\pi\}$. This rendering process employs the differentiable rasterizer of 3DGS~\cite{kerbl20233d}, enabling direct supervision of the multi-view rendering output by the provided ground-truth images $\{\bbx\}$ at both reference and novel viewpoints through image space reconstruction loss.

\noindent \textbf{Condition image tokens.} The reference view $\bbx \in \mathbb{R}^{H \times W \times 3}$ is tokenized using the visual transformer (ViT) model DINO~\cite{dinov2}, which has demonstrated robust feature extraction capabilities in previous large reconstruction models (LRM)~\cite{adobelrm, zou2023triplane}. This process extracts the reference image $\bbx$ into a sequence of tokens $\mathbf{X} \in \mathbb{R}^{K \times C}$, characterized by a length of $K$ and channel dimensions $C$.

\noindent \textbf{Condition camera tokens.} Given the variability of camera poses $\pi_{\text{ref}}$ across the sampled 3D objects during the training phase, it is essential to embed camera features as a condition in our GambaFormer. Following the precedent settings in LRM~\cite{adobelrm}, we construct the camera matrix using 12 parameters that include the rotation and translation components of the camera extrinsics and 4 parameters $[fx, fy, cx, cy]$ representing the camera intrinsics. These parameters are then transformed into a high-dimensional camera embedding $\mathbf{T} \in \mathbb{R}^{C}$ via a multi-layer perceptron (MLP). It is important to note that Gamba does not rely on any canonical pose, and the ground truth $\pi$ is required solely as input during training for multi-view supervision.

% \begin{figure}[t]
% \centering
% \includegraphics[width=0.7\linewidth]{figures/gamba-block-new.pdf}
% \caption{Single Gamba block, where layer normalization (LN), SSM, depth-wise convolution~\cite{chollet2017xception}, and residual connections are employed.}
% \label{fig:gamba_block}
% \end{figure}

\noindent \textbf{Expanding image as 3DGS tokens.} To effectively reconstruct a 3D object using a set of Gaussians, our framework necessitates the prediction of a substantial number of 3D Gaussians. Achieving a sufficient count of Gaussians is essential for accurately fitting a 3D object. Previous methods~\cite{zou2023triplane, agg} have resorted to a two-stage training framework to manage the considerable memory overhead associated with long token sequences, which initially trained a network with capability of predicting up to $L=4096$ Gaussians, and then trained a super-resolution network enhances the resolution of the output from the first stage to $L=16384$. 

In contrast, our GambaFormer architecture obviates this two-stage training paradigm by leveraging the linear complexity of state space models. The Gaussian count is set as $L=16384$ throughout our framework. To construct the 3DGS token sequence, we start by embedding the paired camera pose $\pi_{ref}$ into the reference image $\mathbf{x}_{ref}$ with Plücker rays~\cite{dmv3d} at each image pixel, which are then concatenated into  $\mathbf{s} \in \mathbb{R}^{H \times W \times 9}$. Following this, a large non-overlapping convolution with a kernel size of $p \times p$ is applied to transform $\mathbf{s}$ into a feature map $\mathbf{S} \in \mathbb{R}^{h \times w \times D}$:
\begin{equation}
\mathbf{G} = \text{Scan}(\text{Conv}(\mathbf{s})) + \mathbf{E},
\label{eq:gs_tokens}
\end{equation}
where $h = H / p$ and $w = W / p$. The dimension of each Gamba block is denoted by $D$. We then employ four pre-defined scan orders~\cite{vmamba} to flatten this feature map into 1D sequence of length $L = 4 \times h \times w$. Finally, this sequence is plus with learnable 3D Gaussian Splatting (3DGS) embeddings, $\mathbf{E} \in \mathbb{R}^{L \times D}$, resulting in the formation of 3DGS tokens, $\mathbf{G} \in \mathbb{R}^{L \times D}$, which serve key input to our GambaFormer.

\textbf{Core of the Gamba Block.} The detailed architecture of the Gamba block, compared with the vanilla Mamba block, is illustrated in Figure~\ref{fig:overall_gambaformer}(b). While the Mamba block excels at processing long sequences of tokens, existing variants~\cite{vmamba, visionmamba, pointmamba} have not explored traditional cross-attention mechanisms. Leveraging the unidirectional scan order inherent to Mamba, we aim to utilize this feature for conditional prediction. The Gamba block is composed of a Mamba block, two linear projections, and straightforward $\text{Prepend}$ and $\text{Drop}$ operations:
\begin{equation}
\begin{aligned}
\mathbf{H}_{n} &= M_{n}(\text{Prepend}(\mathbf{P}_{c}^{n} \mathbf{T}, \mathbf{P}_{x}^n \mathbf{X}), \mathbf{G}_{n-1}), \\
\mathbf{G}_{n} &= \text{Drop}(\mathbf{H}_{n}, \text{Index}(\mathbf{P}_{c}^n \mathbf{T}, \mathbf{P}_{x}^n \mathbf{X})),
\end{aligned}
\end{equation}
where $M_{n}$ represents the $n$-th vanilla Mamba block. The $\mathbf{P}_c^n \in \mathbb{R}^{D \times C}$ and $\mathbf{P}_x^n \in \mathbb{R}^{D \times C}$ are learnable linear projections for camera and image tokens, respectively, in the $n$-th layer. The operation $\text{Prepend}$ involves adding projected camera embeddings and image tokens to the beginning of the sequence before processing through the hidden 3DGS features $\mathbf{G}_{n - 1}$ in each layer. Conversely, $\text{Drop}$ removes the earlier prepended tokens from the output $\mathbf{H}_{n}$, based on their indexed positions.

\noindent \textbf{Gaussian Decoder.} With $N$ stacked Gamba blocks, our GambaFormer adeptly extracts hidden features for each 3DGS token condition on the reference image. A sophisticated Gaussian Decoder then decodes the attributes of each Gaussian $\mathcal{G}_j$. Initially, the output $\mathbf{G}_{N - 1}$ from the GambaFormer is input into a shallow MLP, encoding the 3DGS tokens as $\mathbf{Z} = \Phi_{\theta}$, where $\Phi_{\theta}$ represents the MLP with learnable parameters $\theta$, and $\mathbf{Z} \in \mathbb{R}^{L \times D}$. Subsequently, separate linear projections are applied to predict each attribute. To precisely predict the $N$ central positions $\mu_j$, we discretize the coordinate space $\mu_j \in [-0.5, 0.5]^3$ into 21 uniformly spaced coordinate points $c_j$. A linear projection $\mathbf{W} \in \mathbb{R}^{21 \times D}$ then maps $\mathbf{Z}$ to $\mathbf{Q} \in \mathbb{R}^{N \times 21}$, and the predicted coordinate can be denoted as:
\begin{equation}
P(Q_{ij}) = \frac{e^{Q_{ij}}}{\sum_{k=1}^{21} e^{Q_{ik}}}, \quad y_i = \sum_{j=1}^{21} P(Q_{ij}) \times c_j, \quad \text{for } c_j \in \{-0.5, -0.45, \ldots, 0.45, 0.5\}
\end{equation}
Here, for simplicity, the formulation only considers one axis. $P(Q_{ij})$ represents the softmax probability of the $j$-th coordinate point for the $i$-th token, and $y_i$ denotes the predicted coordinate for the $i$-th token. Opacity $\alpha_{j}$ is derived using a linear projection followed by a $\text{Sigmoid}$ activation. Scale $r_{j}$ is predicted using a linear projection with a $\text{Softplus}$ activation without constrain. 
% Additionally, rotation $q_{j}$ is fixed as $(1, 0, 0, 0)$ to stabilize training. 
Additionally, only the $0$-th order of spherical harmonics $c_{i}$, constrained within the RGB space, is predicted.

\subsection{Robust Amortized Training.}
\label{subsec:robust-train}
\noindent\textbf{Gaussian Parameter Constraints.}
Learning accurate 3D positions of Gaussians from a single image presents significant challenges due to the limited geometric information available from a single viewpoint. 
Prior works~\cite{agg, zou2023triplane} has employed point clouds, sampled from ground-truth meshes, as supervision during the initial training phase to prevent model collapse. 
\begin{wrapfigure}{r}{0.42\textwidth}
% \vspace{-4mm}
  \centering
  \includegraphics[width=0.4\textwidth]{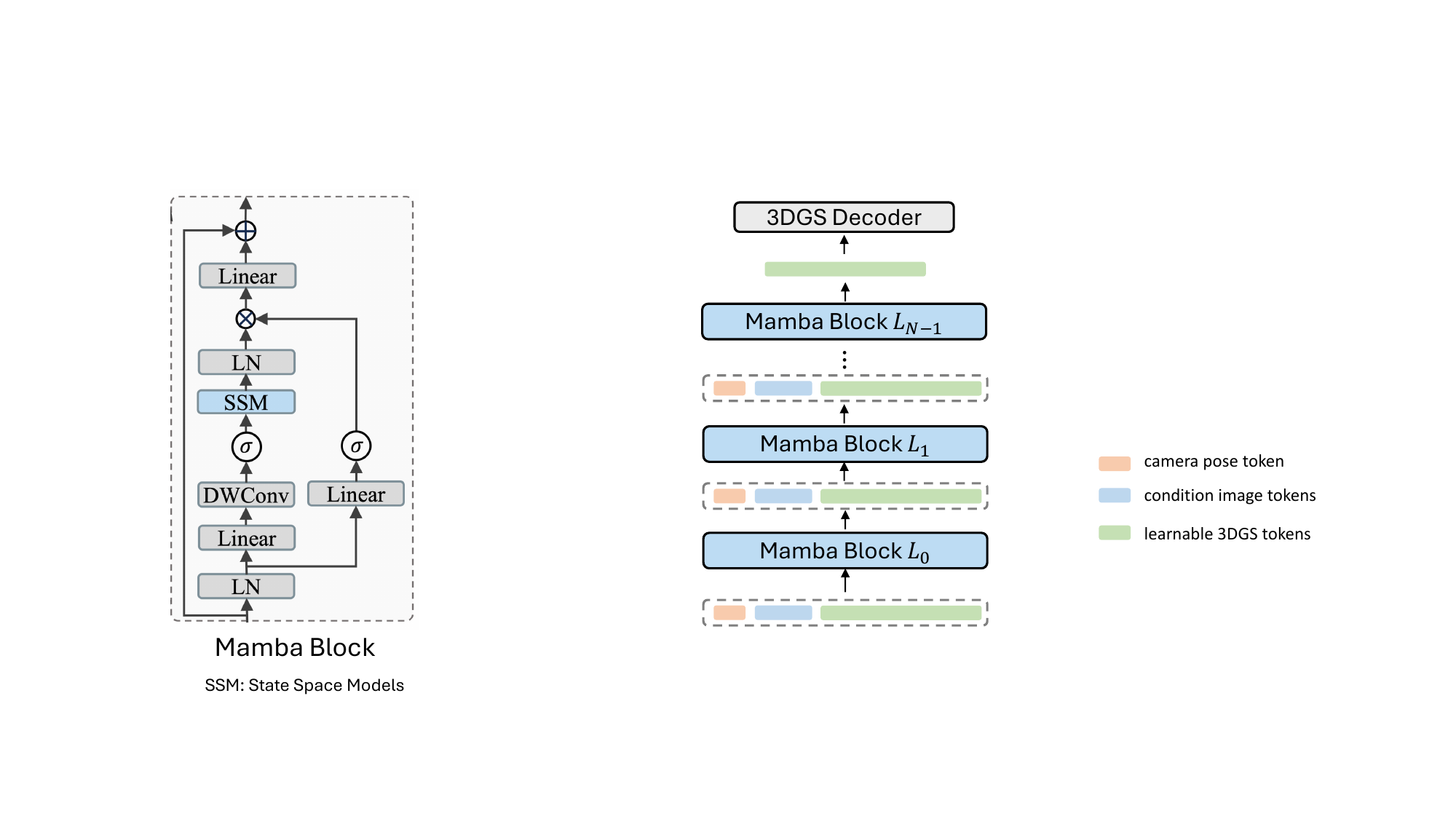}
  \caption{\textbf{Radial polygon mask}. Object masks are divided into polygon masks by 2D ray casting from the image center to the contours.}
  \label{fig:ray_mask}
  \vspace{-4mm}
\end{wrapfigure}
This form of supervision, however, impedes both the end-to-end training and the scalability of large reconstruction models.

To address these issues, we introduce a \textit{radial mask constraint}, inspired by the notion that images from multiple viewpoints can depict the occupied 3D space of an object. As illustrated in Figure~\ref{fig:ray_mask}, the view mask $\bbx_{mask}$ is first discretized into a distance field for fast approximation. Using 2D ray casting from the image center to the mask contour, we obtain a set of radial contour distance fields $\mathbf{v} \in \mathbb{R}^{U}$, where $U$ denotes the number of rays. If the projected 2D center of a Gaussian falls outside these contours, an explicit loss is applied to correct its position:
\begin{equation}
\mathcal{L}_{\text{rdist}} = \mathcal{L}_{\text{MSE}}\left(\text{Interp}(v^k, v^{k+1}), d_j\right), \quad \text{where } d_j = \sqrt{(x_j - c_x)^2 + (y_j - c_y)^2}
\label{eq:radial_mask}
\end{equation}
Here, $d_j$ represents the distance from the 2D projected center of Gaussian $G_j$ to the image center $(c_x, c_y)$, and $\text{Interp}(v^k, v^{k+1})$ denotes the linearly interpolated ground truth distance between two adjacent rays $v^k$ and $v^{k + 1}$, based on the radial angle of the projected 2D Gaussians.

This image-based constraint allows us to eliminate the need for explicit 3D supervision previously required in single-view reconstruction models. By imposing such constraints on the projected 2D Gaussians, the positions of the Gaussians can rapidly converge to rough 3D shapes.

\noindent\textbf{Training Objective.}
Utilizing the efficient tiled rasterizer for 3D Gaussians~\cite{kerbl20233d}, Gamba is trained end-to-end with image-space reconstruction loss across both reference and novel views. The training loss function is formulated as follows:
\begin{equation}
\begin{aligned}
\mathcal{L}_{train} = \frac{1}{V + 1}\sum_{i=0}^{V} &\mathcal{L}_{\text{MSE}}(\hat{v}_{i}^{\text{rgb}}, v_{i}^{\text{rgb}}) + \lambda_{\text{mask}}\mathcal{L}_{\text{MSE}}(\hat{v}_{i}^{\alpha}, v_{i}^{\alpha}) \\
&+ \lambda_{\text{LPIPS}}\mathcal{L}_{\text{LPIPS}}(\hat{v}_{i}^{\text{rgb}}, v_{i}^{\text{rgb}}) + \lambda_{\text{rdist}}\mathcal{L}_{\text{rdist}},
\end{aligned}
\label{eq:loss_train}
\end{equation}
where $\hat{v}_i^{\text{rgb}}$ and $\hat{v}_i^{\alpha}$ denote the predicted RGB image and alpha mask, respectively, rendered from the predicted 3D Gaussians $\{\mathcal{G}\}$, while ${v}_i^{\text{rgb}}$ and ${v}_i^{\alpha}$ represent the corresponding ground truth. $\mathcal{L}_{\text{LPIPS}}$ encompasses the VGG-based perceptual loss for image fidelity~\cite{zhang2018unreasonable}, and $\mathcal{L}_{\text{rdist}}$ is the radial mask constraint loss defined in Eq.~\eqref{eq:radial_mask}. Additionally, the $\lambda_{mask}$, $\lambda_{\text{LIPIPS}}$ and $\lambda_{\text{rdist}}$ are balancing factors. These losses are applied across $V + 1$ viewpoints, including the input reference viewpoint and $V$ novel viewpoints to supervise geometry and texture traininig together.

Inspired by the coarse-to-fine optimization strategies employed in SDS-based image-to-3D reconstructions~\cite{poole2022dreamfusion, dtc123, qian2023magic123}, we progressively increase the number of views from $2$ to $6$ during training. This strategy not only reduces the computational load but also enhances the model’s robustness by allowing gradual adaptation from geometry to detailed textures. 
% Additionally, we have developed specific view sampling strategies to robustly reinforce geometric learning (refer to supplementary material Sec.~\ref{sec:robust_train} for details).

% \subsection{Inference}
% After training on the large scale 3D datasets~\cite{xx}, 
\label{inference}
\section{Experiments}

% \subsection{Datasets}

\subsection{Implementation Details}

\noindent \textbf{Datasets.}
Following previous work~\cite{liu2023one, zou2023triplane}, we utilized a filtered LVIS subset of the Objaverse dataset~\cite{deitke2023objaverse} for pre-training Gamba, which comprises around 40k 3D models across 1,156 categories. We filtered this subset by intersecting it with 3D objects rendered in the G-buffer Objaverse~\cite{richdreamer}, resulting in a final training set of approximately 20k high-quality 3D objects. Additionally, to improve model generalization, the camera poses for rendered objects during training are normalized to a unit distance~\cite{tang2024lgm}. For evaluation, our trained model was qualitatively assessed using web images. Quantitatively, we conducted comparisons on the Google Scanned Object (GSO) dataset~\cite{downs2022google}, selecting totally 60 3D objects and rendering a single view of each at a spatial resolution of $512 \times 512$ for comprehensive evaluation.

\noindent \textbf{Network Architecture.} The GambaFormer architecture comprises $N=14$ Gamba blocks, each with hidden dimensions $D = 512$. For the condition image tokens, we employ the pretrained DINO v2 model~\cite{dinov2} as our image tokenizer, which extracts semantic feature tokens of length $K=576$ from the reference image. To construct the 3DGS token sequence, the reference image, with a spatial resolution of $512 \times 512$, is initially processed with a convolution kernel with $p = 8$. This step tokenizes the image into 4096 tokens. Subsequently, four pre-defined scan orders~\cite{vmamba} are applied sequentially to expand this into a token sequence of length $L=16384$. The 3DGS embeddings $\mathbf{E}$ are learnable positional embeddings and correspond to 16384 3D Gaussians, matching the length of the above tokens. The Gaussian Decoder employs a straightforward MLP architecture with a single hidden layer, complemented by separate linear projections for each attribute of the Gaussians.

\noindent \textbf{Pre-training.} Gamba is trained on 16 NVIDIA A100 (80G) GPUs with a batch size of 256 over approximately 40 hours for totally 400 epochs. We employ the AdamW optimizer~\cite{loshchilov2017decoupled} with an initial learning rate of 1e-3 and a weight decay of 0.05. Gradient clipping of value $1.0$ was implemented to maintain the $L_2$ norm of gradients. The loss weight settings are as follows: $\lambda_{\text{mask}} = 1$ for the mask loss and $\lambda_{\text{LPIPS}} = 0.5$ for LPIPS~\cite{zhang2018unreasonable} loss. Additionally, the loss weight $\lambda_{\text{rdist}}$ for the radial mask constraints is initially set at 0.1 and is gradually decayed to 0 over the first 10 epochs, aiming to ensure that the predicted positions of 3D Gaussians converge within a reasonable range in 3D space.

\noindent \textbf{Inference.} During inference, Gamba only takes an arbitary RGB image as input, where the foreground object is segmented using a pre-trained segmentation model~\cite{kirillov2023segany} and subsequently recentered. Gamba employs a default camera pose with both zero elevation and azimuth, which is used to produce camera tokens and Plücker ray inputs. Gamba efficiently predicts $16,384$ Gaussians for a single 3D object in a feed-forward manner. Remarkably, this process requires only about \textit{8 GB} of GPU memory and completes in less than \textit{0.05 second} on a single NVIDIA A100 (80G) GPU, making it well-suited for online deployment scenarios.

\begin{figure}[t]
\centering
\includegraphics[width=\linewidth]{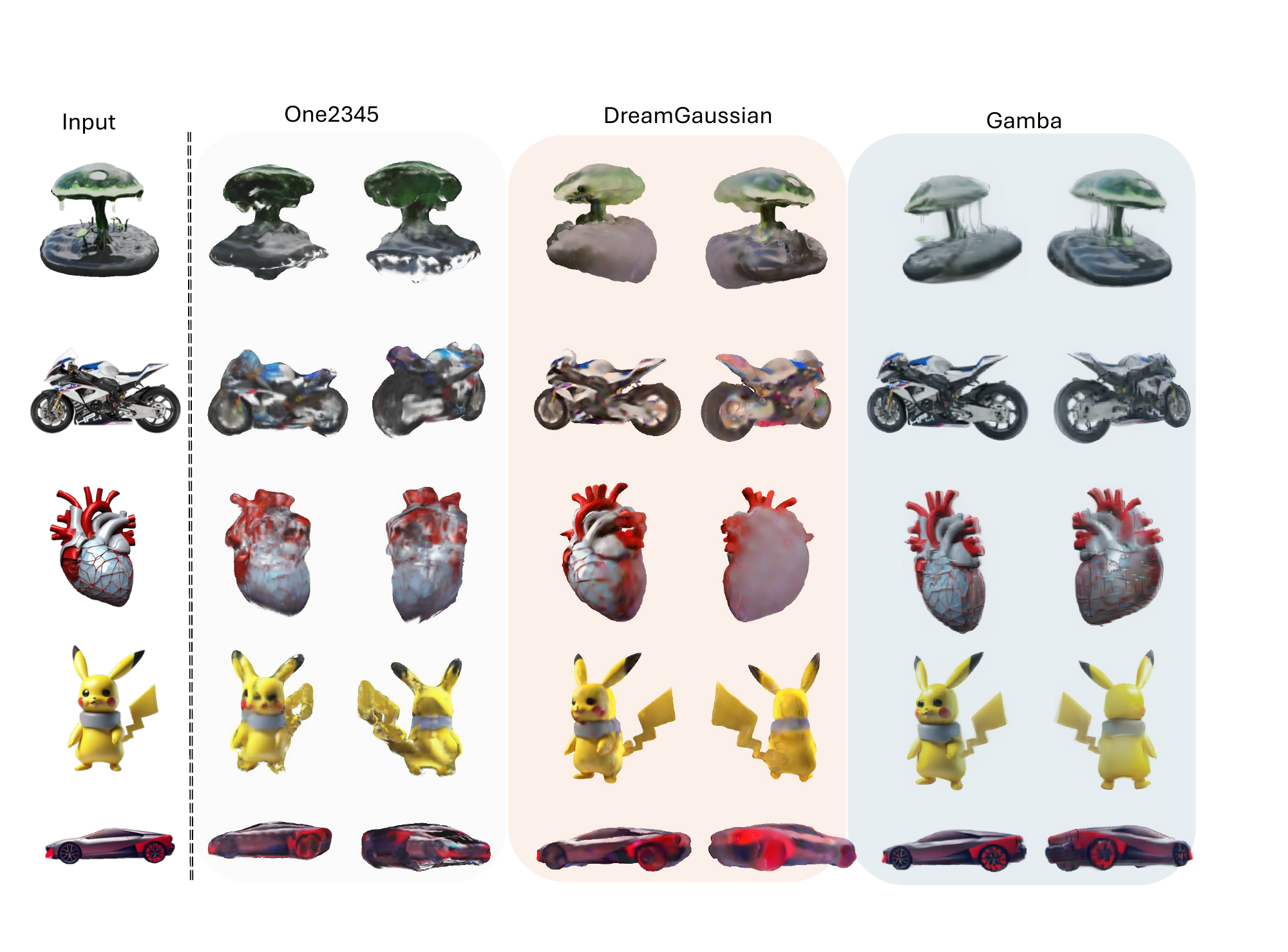}
\caption{Qualitative Comparison with large reconstruction models.}
\label{fig:exp1}
\vspace{-4mm}
\end{figure}

\subsection{Experimental Protocol} 

% \noindent \textbf{Baseline.} We compare our Gamba with previous state-of-the-art single-view reconstruction methods. The most related stream is large reconstruction models, trained with a substantial rendered multi-view images and only need to feed-forward in inference. LRM~\cite{adobelrm} predict Tri-Plane representation from a single image in a transformer-based architecture. Triplane-Meets-Gaussian (TGS) also take advantage the fast rendering of 3DGS, it first predict $16384$ points cloud of a 3D object, then other attributes of Gaussians are predicted in another transformer network. Another stream of single view 3D reconstruction method is SDS-based~\cite{poole2022dreamfusion}. These methods utilize the multi-view diffusion model Zero-1-to-3~\cite{liu2023zero} trained on Objaverse-XL~\cite{ObjaverseXL}, which can generate multi-view images condition on single image and relative camera poses. DreamGaussian~\cite{tang2023dreamgaussian} is the fastest method among them, that can generate 3D object in 1 minute. One-2-3-45 circumvente costly optimization by leveraging a generalizable SparseNeuS~\cite{wang2021neus,wang2023neus2} to directly predict SDF from generated multi-view images. Note that AGG~\cite{agg} has not been included for comparison, as no code and results has been publicly released. 

\noindent \textbf{Baseline.} We benchmark Gamba against previous single-view reconstruction methods, particularly those in the stream of large reconstruction models. These models are typically trained on a large number of rendered multi-view images and are designed for efficient feed-forward inference. The first work, LRM~\cite{adobelrm} utilizes a transformer-based architecture to predict a Tri-Plane representation from a single image. Triplane-Meets-Gaussian (TGS) combines the fast rendering capabilities of 3DGS by initially predicting a point cloud of $16,384$ for a 3D object, followed by predicting other Gaussian attributes using another transformer network. In contrast, another stream is SDS-based methods with iterative optimization like DreamGaussian~\cite{tang2023dreamgaussian}. These approaches leverage a multi-view diffusion model, Zero-1-to-3~\cite{liu2023zero}, trained on the Objaverse-XL dataset~\cite{ObjaverseXL}, to produce multi-view images conditioned on a single image and relative camera poses. Another notable method, One-2-3-45, bypasses the need for costly optimization by utilizing a generalizable SparseNeuS~\cite{wang2021neus,wang2023neus2} to directly predict Signed Distance Functions (SDF) from generated multi-view images. 
% It is important to note that AGG~\cite{agg} has been excluded from our comparison due to the unavailability of both its code and published results.

\begin{figure}[h]
\centering
\includegraphics[width=\linewidth]{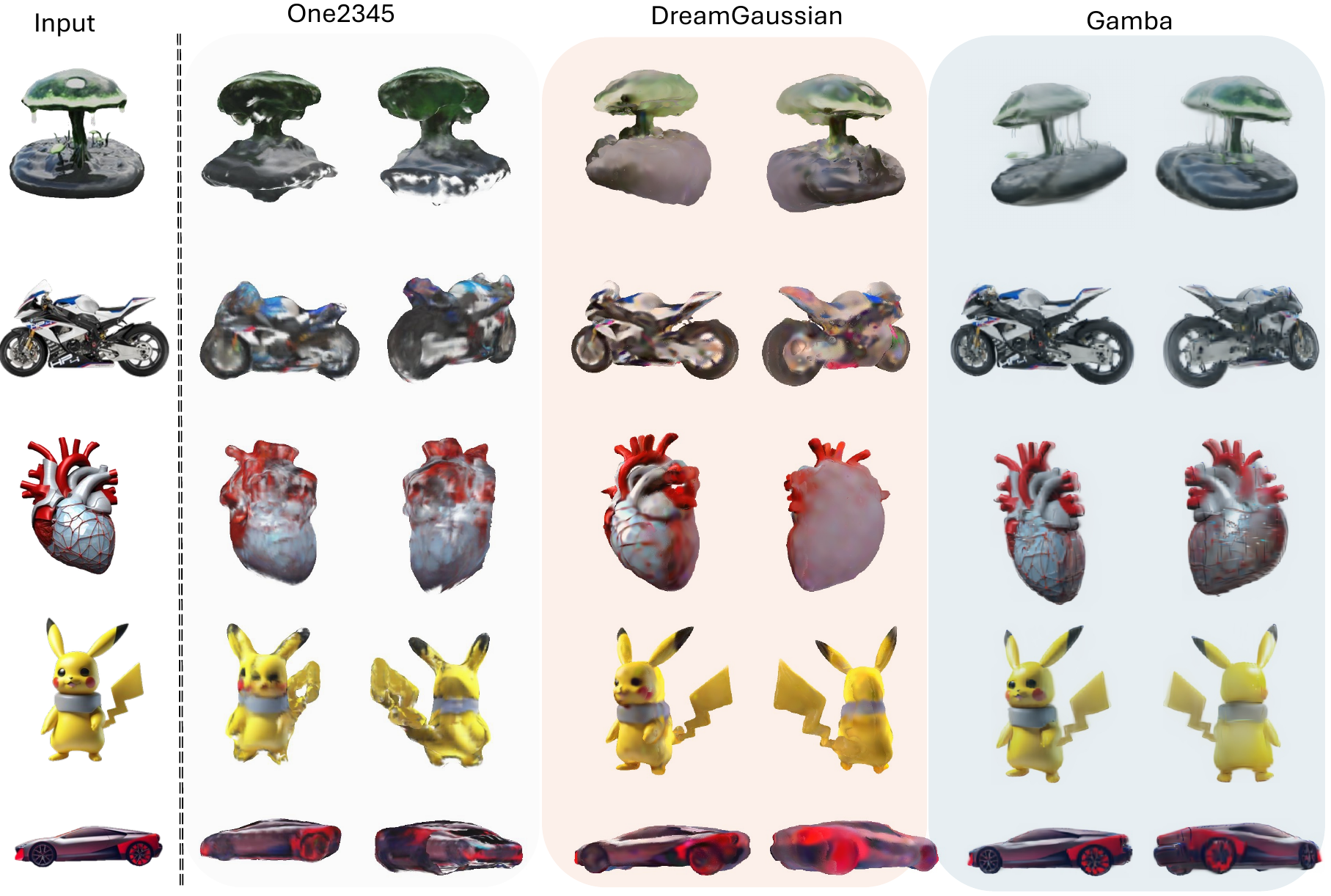}
\caption{Comparison with Zero-1-to-3~\cite{liu2023zero} based single-view 3D reconstruction methods, including feed-forward only method One-2-3-45~\cite{liu2023one} and optimization-based DreamGaussian~\cite{tang2023dreamgaussian}.}
\label{fig:exp2}
\vspace{-4mm}
\end{figure}

\noindent\textbf{Qualitative Comparisons.}
Figures.~\ref{fig:exp1} and~\ref{fig:exp2} demonstrate Gamba's capability to maintain reasonable geometry and plausible textures in reconstructing various 3D objects. In contrast, reconstructions by most baseline methods suffer from multi-view inconsistency and geometric distortion. Despite LRM being trained on a dataset that is $50 \times$ larger than ours, it frequently exhibits warped geometries (row 1, 2, 5 in Figure~\ref{fig:exp1}). Compared to TGS, which also employs 3DGS for representation, Gamba consistently delivers better texture reconstruction (especially row 1, 3, 4 in Fig.~\ref{fig:exp1}). Fig.~\ref{fig:exp2} further highlights that, while both One-2-3-45 and DreamGaussian leverage the advanced Zero-1-to-3-XL model~\cite{liu2023zero, ObjaverseXL}, their reconstruction still exhibit artifacts with multi-view inconsistency and geometric distortion (rows 1 and 3) alongside blurred textures (rows 2, 3, and 4). This comparative analysis underscores Gamba's robustness and superior performance in single-view 3D reconstruction.

% \vspace{4mm}

\noindent\textbf{Quantitative Comparisons.}
Following prior works, we adopt PSNR, SSIM, and LPIPS metrics to evaluate the quality of novel view synthesis. Additionally, we employ Chamfer Distance and Volume IoU between ground truth and reconstruction to quality of geometry reconstruction. Gamba is compared against 2 SDS-based methods: Zero-1-to-3~\cite{liu2023zero} and DreamGaussian~\cite{tang2023dreamgaussian}, as well as 5 feed-forward only methods: One-2-3-45~\cite{liu2023one}, the point cloud diffusion model Point-E~\cite{nichol2022point}, the NerF parameter diffusion model Shap-E~\cite{shap-e}, the pioneering large reconstruction model LRM~\cite{adobelrm}, and TGS which integrates LRM with 3DGS~\cite{zou2023triplane}. The results, presented in Table~\ref{table:Image-to-3D}, indicate that Gamba is competitive in both texture and geometry reconstruction.

\begin{table}[h!]
\vspace{-2mm}
\centering
\caption{\textbf{Quantitative results.} We evaluate novel view synthesis in terms of PSNR$\uparrow$/ SSIM$\uparrow$/LPIPS$\downarrow$ and geometry reconstrution in terms of Chamfer Distance$\downarrow$/Volume IoU$\uparrow$.}
\resizebox{0.85\linewidth}{!}{
\begin{tabular}{c|ccc|cc|c}
\hline
              & PSNR $\uparrow$ & SSIM  $\uparrow$ & LPIPS  $\downarrow$ & Chamfer Dist.  $\downarrow$ & Volume IoU  $\uparrow$ & Time  $\downarrow$ \\ \hline
One-2-3-45~\cite{liu2023one}    & 14.93           & 0.72             & 0.28                & 0.0683                      & 0.3485                 & 40s                \\
Point-E~\cite{nichol2022point}       & -               & -                & -                   & 0.0515                      & 0.2643                 & 78s                \\
Shap-E~\cite{shap-e}       & -               & -                & -                   & 0.0579                      & 0.3228                 & 27s                \\
Zero-1-to-3~\cite{liu2023zero}   & 18.28           & 0.76             & 0.19                & 0.0385                      & 0.3786                 & 1800s              \\
DreamGaussian~\cite{tang2023dreamgaussian} & 21.65           & 0.82             & 0.14                & 0.0341                      & 0.3615                 & 70 s               \\
LRM~\cite{adobelrm}           & 21.35           & 0.82             & 0.15                & 0.0325                      & 0.3872                 & 0.5s               \\
TGS~\cite{zou2023triplane}           & 22.68           & 0.85             & 0.12                & 0.0257                      & 0.4121                 & 0.2s               \\ \hline
\textbf{Ours} & \textbf{24.74}  & \textbf{0.91}    & \textbf{0.08}       & \textbf{0.0232}             & \textbf{0.4289}        & \textbf{0.05s}     \\ \hline
\end{tabular}

}
\label{table:Image-to-3D}
\vspace{-2mm}
\end{table}

\noindent{\textbf{Inference Runtime}}
We showcase the inference runtime required to generate a 3D asset in Table~\ref{table:Image-to-3D}, where the timing is recorded using the default hyper-parameters for each method on a single NVIDIA A100 GPU (80G). Remarkably, our Gamba outperforms optimization-based approaches like Zero-1-to-3~\cite{liu2023zero} in terms of speed, being several orders of magnitude faster than those optimization-based methods and surpass other feed-forward models as well, thanks to the efficient backbone design.

\section{Ablation and Discussion}
\vspace{-2mm}
In ablation studies, all experiments is conducted on a randomly selected subset from G-buffer~\cite{richdreamer} Objaverse, around totally 10k training data, and $500$ objects in this subset are left for evaluation. And all models are trained 100 epochs only for evaluation.

\noindent\textbf{Q1: }\emph{\textbf{What impacts performance of Gamba in terms of component-wise contributions?}} We discarded each core component of Gamba to validate its component-wise effectiveness. The results are described in Table~\ref{tab:component_ablation} by comparing their predicted views with ground-truth multi-view image set on the evaluation set. 

\vspace{-2mm}

\noindent\textbf{A1: } 
In an end-to-end, multi-component pipeline, we observed that the exclusion of any component from Gamba resulted in a significant degradation in performance. Specifically, we first remove the loss term of the radial mask constraint by setting $\lambda_{\text{rdist}} = 0$ in Eq.~\eqref{eq:loss_train} for the ``w/o radial mask constraint". We find that the training is prone to collapse, i.e., the 3D position of predicted Gaussians confines to a small sphere, after which all predicted opacities $\alpha$ become 0, leading to Gaussians becoming invisible in 2D rendering outputs. This removal thus produces a catastrophic performance drop; the evaluation PSNR is only 12.72. Furthermore, in comparison to prior works such as AGG~\cite{agg} and Triplane-Meets-Gaussian~\cite{zou2023triplane}, which exclusively utilized learnable positional embeddings as 3D tokens, our model was also evaluated under this variant. The ``w/o additive 3DGS tokens" scenario means taking the 3DGS tokens solely from learnable embeddings, i.e., setting $\mathbf{G} = \mathbf{E}$ in Eq.~\eqref{eq:gs_tokens}; the evaluation PSNR degraded to 20.35 in quantitative evaluation. In qualitative comparison, we find that the reconstruction after this modification tends to produce over-smoothed texture, and the generalization of the model to different objects is also degraded. Finally, we removed the prepending operation of the conditional camera tokens and image tokens. The evaluation results, shown in ``w/o Prepending," reveal that removing this component leads to a 1.24 dB drop in terms of PSNR, which demonstrates the necessity of this prepending operation.

\begin{table}[h]
\vspace{-4mm}
\caption{Ablation Studies of component-wise contribution.}
\centering
\begin{tabular}{c|ccc}
\hline
Model variants            & PSNR $\uparrow$ & SSIM  $\uparrow$ & LPIPS  $\downarrow$ \\ \hline
w/o radial mask constraint & 12.72           & 0.58             & 0.47                \\
w/o additive  3DGS tokens & 20.35           & 0.79             & 0.16                \\
w/o Prepending            & 22.57           & 0.85             & 0.10                \\
Full Model                & 23.81           & 0.88             & 0.12                \\ \hline
\end{tabular}
\label{tab:component_ablation}
\end{table}

\noindent\textbf{Q2:} \emph{\textbf{Why don't we apply more Gaussians?}} Mamba~\cite{mamba} exhibits linear computational complexity and memory consumption with respect to token length. We empirically validated that $L=16384$ Gaussians are sufficient for amortized 3D reconstruction in our end-to-end training framework.
\vspace{-2mm}

% \begin{figure}[h!]
% \centering
% \includegraphics[width=0.5\linewidth]{figures/complexity.pdf}
% \caption{Mamba vs Transformer memory consumption comparison over token length. }
% \label{fig:mamba_transformer}
% \end{figure}

\begin{figure}[h!]
\centering
\includegraphics[width=0.5\linewidth]{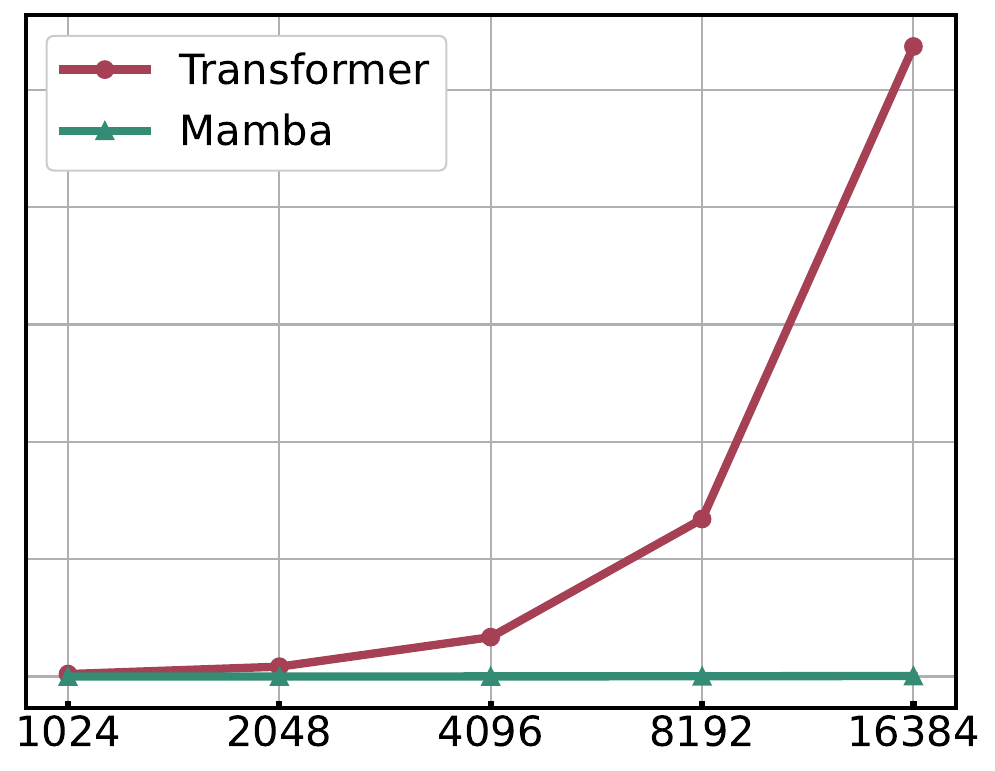}
\caption{Mamba vs Transformer memory consumption comparison over token length. }
\label{fig:mamba_transformer}
\end{figure}

\noindent{\textbf{A2}}: We illustrate the memory consumption of Mamba and Transformer in Figure~\ref{fig:mamba_transformer}. Existing 3DGS-based amortized 3D reconstruction models~\cite{agg, zou2023triplane}, which are all built on transformer architectures, have a maximum of $4096$ 3D Gaussians due to the $O(N^2)$ memory consumption of transformers. These models employ local attention or 1D-CNN to predict offsets over the $4096$ Gaussians to achieve a higher resolution, such as $16384$ Gaussians from $4096$ base Gaussians. However, these models are trained in two stages, which limits the performance of the higher resolution Gaussians by the capabilities of the first-stage model. In contrast, our method, which is trained end-to-end, directly applies $16384$ Gaussians in a single forward. Additionally, we quantitatively reconstructed $500$ objects using original 3D Gaussian Splatting with multi-view images and statistically analyzed their total Gaussians after iterative reconstruction. We found that $87\%$ of the 3D objects maintained Gaussian counts between $10000$ and $20000$. Thus, using $L=16384$ for Gaussian numbers is sufficient in our Gamba framework.

\noindent\textbf{Q3:} \emph{\textbf{Why do we not use Tri-plane representation?}} Almost all previous amortized single view reconstruction models integrate Tri-plane representation to encode textures or the geometry of 3D objects. We also explored adding the Tri-plane representation to our Gamba model for ablation study.

\begin{figure}[h!]
    
    \centering
    \begin{subfigure}[b]{0.55\textwidth}
        \includegraphics[width=\linewidth]{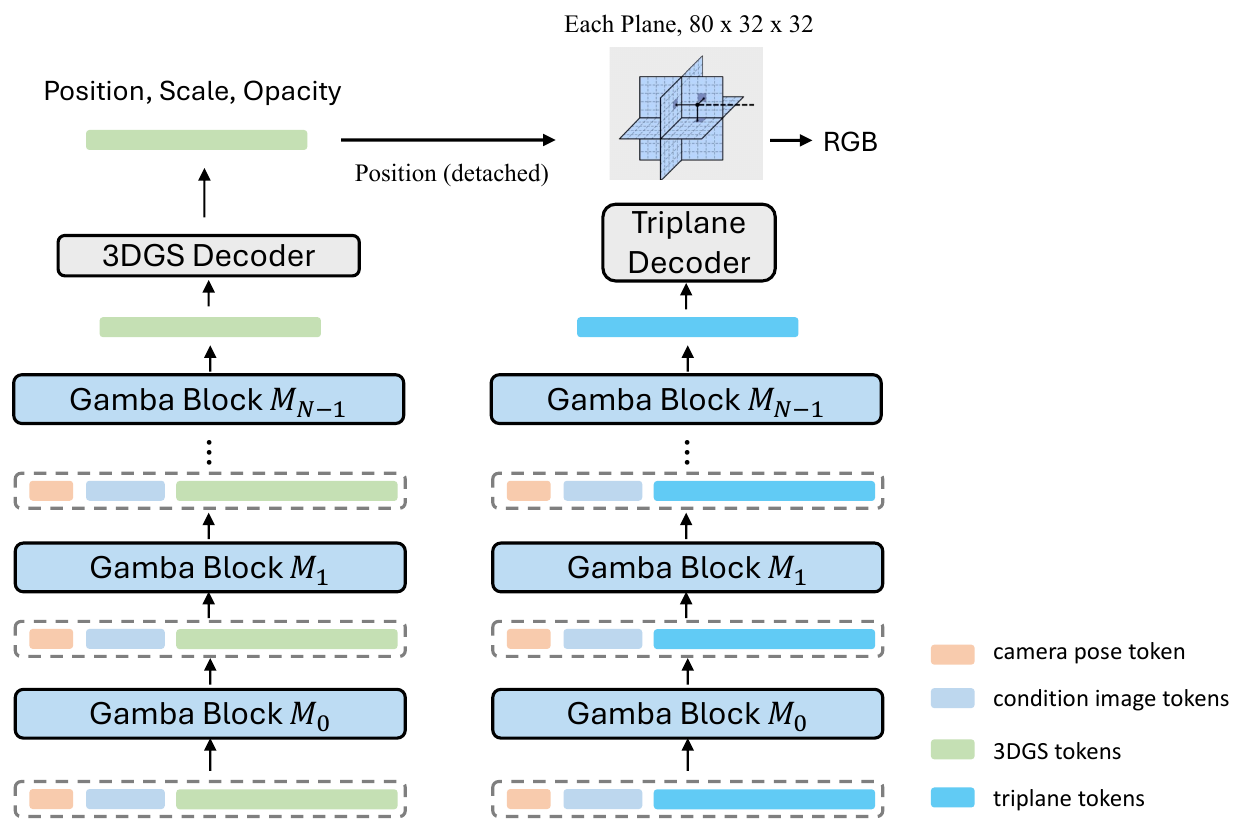}
        \caption{Architecture of Gamba meets Tri-plane}
    \end{subfigure}
    \hfill
    \begin{subfigure}[b]{0.42\textwidth}
        \includegraphics[width=\linewidth]{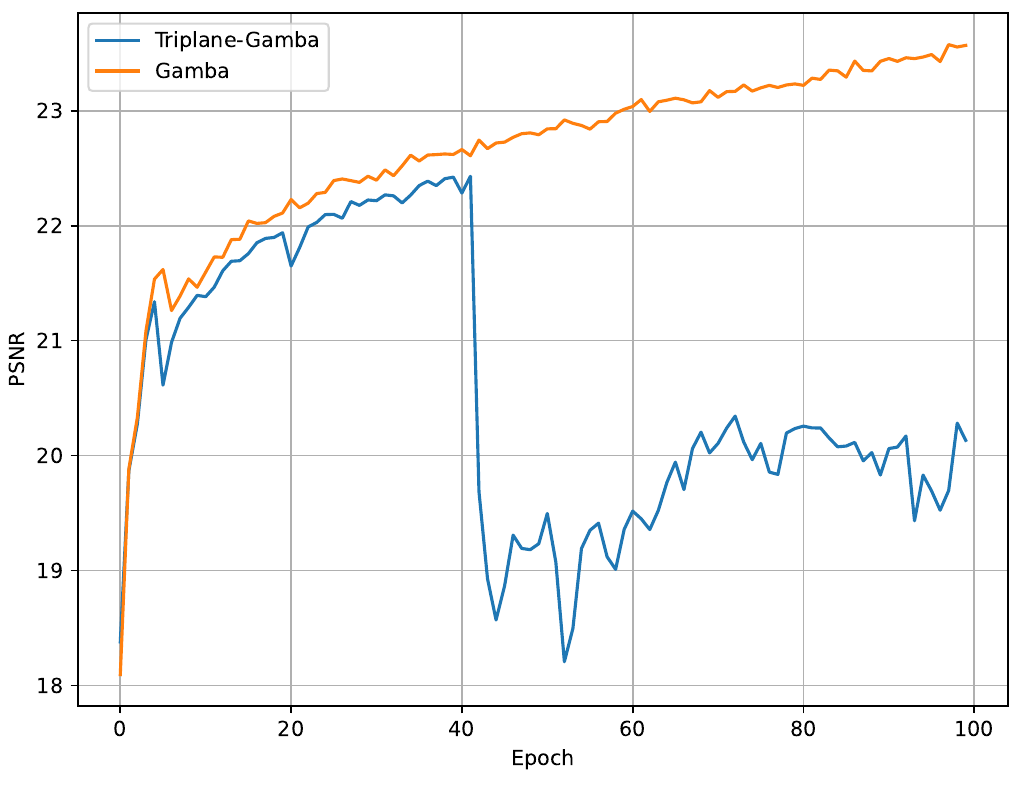}
        \caption{Evaluation PSNR on G-buffer Objaverse.}
    \end{subfigure}
    \caption{Ablation study of Tri-plane meets Gamba.}
    \label{fig:triplane_ablation}
\end{figure}

\noindent{\textbf{A3}}: In Figure~\ref{fig:triplane_ablation} (a), we show that our Tri-Plane meets-Gaussian structure. A Siamese GambaFormer architecture is adopted where the left GambaFormer predicts positions $\mu_j$, opacity $\alpha_j$, and scales $r_j$. Similarly, the right branch adopts a methodology as described in Triplane-Meets-Gaussian~\cite{zou2023triplane}. Here, the predicted positions of Gaussians $\mu_j$ serve as queries to extract hidden features from the predicted hidden Tri-Plane features. Subsequently, these queried features from three planes are concatenated to decode the color $c_j$ of each Gaussian. Ultimately, the prediction results from these two branches are combined and rendered into multi-view images for supervision, employing the same supervision as in Eq.~\eqref{eq:loss_train} during training. The evaluation of PSNR on the selected G-buffer Objaverse subset is shown in Figure~\ref{fig:triplane_ablation} (b). Despite having doubled parameters, the PSNR of ``Triplane-Gamba'' is significantly worse than our base version, and its PSNR sharply degrades at the 40th epoch. This ablation study demonstrates that the Tri-Plane representation is not necessary in our Gamba Framework.

\noindent\textbf{Q4:} \emph{\textbf{Are there any other alternatives to construct the $L$ 3DGS tokens?}} In our initial experiment (Q1), we explored constructing the $L$ 3DGS tokens solely using learnable positional embeddings $\mathbf{E}$. This approach resulted in a significant performance drop, indicating the insufficiency of relying purely on embeddings for constructing 3DGS tokens. In the current methodology, as detailed in Eq.~\eqref{eq:gs_tokens}, we expand the tokenized images by scanning them four times to produce $L=16384$ tokens. This method, while simple yet effective, raises concerns that the reconstructed 3D objects might merely duplicate the conditional image. To verify this concerns, we explore to construct the $L$ 3DGS tokens from a mixed strategy.

\begin{figure}[h]
\centering
\includegraphics[width=0.7\linewidth]{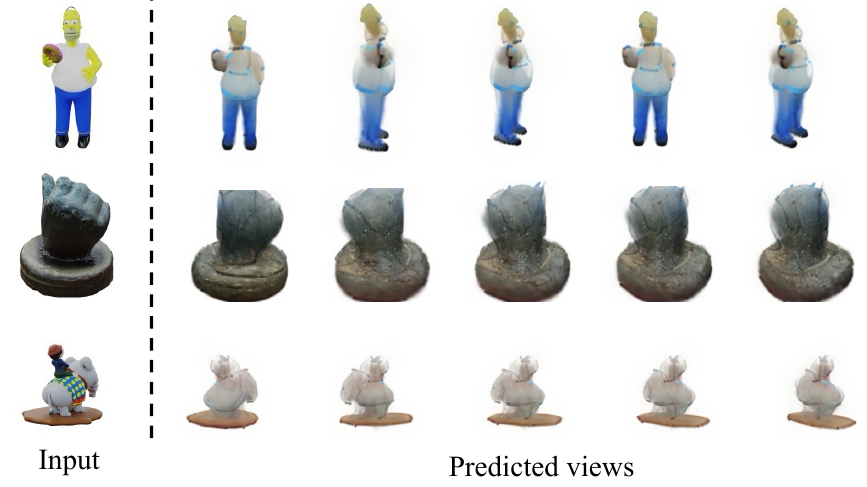}
\caption{Qualitative evaluation of mixed 3D GS tokens. Predicted 3D objects on 2D views rendering appears to be blurred even at the reference viewpoint.}
\label{fig:samba-ablation}
\end{figure}

\noindent{\textbf{A4}}: Specifically, we initially construct $1024$ tokens using a convolution with a kernel size of $p=8$. These tokens are then expanded to $4096$ through four scans, as specified in Eq.~\eqref{eq:gs_tokens}. The remaining $12,288$ tokens are constructed directly from pure positional embeddings $\mathbf{E}$. This method's qualitative results are illustrated in Figure~\ref{fig:samba-ablation}, where the rendered 2D views from the predicted 3D Gaussians exhibit noticeable blurring across all views, including the reference view. This degradation in quality relative to our baseline methods indicates that relying solely on positional embeddings can lead to significant artifacts, affecting both geometry and texture clarity. Our analysis suggests that the difficulty in achieving convergence with positional embeddings during training, combined with their limited capacity to encapsulate relevant information from the conditional image. Conversely, embedding the image directly onto each 3DGS token through four different scan order helps the model progressively predict finer details of the 3D object, aligning more closely with the unidirectional scan on the input 3DGS tokens.

% \noindent{\textbf{More ablations and discussions.}} We conduct more ablation studies and discussions in suppplementary material Sec.~\ref{sec:more_ablation}, including intergrating Tri-Plane in Gamba and the 3DGS tokens in Gamba.
\vspace{-4mm}
\section{Conclusion}
In this work, we present Gamba, the first end-to-end trained, amortized 3D reconstruction model from single-view image. Our proposed Gamba, different from previous methods reliant on SDS and NeRF, marries 3D Gaussian splatting and Mamba to address the challenges of high memory requirements and heavy rendering process. Our key insight is the relationship between the 3DGS generation process and the sequential mechanism of Mamba. Additionally, Gamba integrates several techniques for training stability. Through extensive qualitative comparisons and quantitative evaluations, we show that our Gamba is promising and competitive with several orders of magnitude speedup in single-view 3D reconstruction.

% \section{Limitations}

% \noindent\textbf{{Future Direction}}

% \clearpage
% \clearpage
% \setcounter{page}{1}
\maketitlesupplementary
% \section{More Ablation and Discussion}
\label{sec:more_ablation}

% triplane or not 
% \textbf{why we dont use Triplane representation with 3DGS.}

% \section{More training details}
% \label{sec:robust_train}

% \noindent
% \textbf{Uniform random view sampling.}

% \textbf{Progressive view supervision.}

% \section{Justification of camera pose in the GambaFormer}

% \noindent\textbf{Data Augmentation.} Gamba primarily emphasizes reconstructing foreground objects, rather than modeling the background. Although our training data consist of single images with pure color backgrounds, we have observed that the 2D renderings during inference often present cluttered backgrounds. To avoid over-fitting this pattern, we implement a random color strategy for background generation. Furthermore, we employ a semantic-aware filter~\cite{yi2022identifying} based on the CLIP similarity metric~\cite{radford2021learning}, to select the canonical pose as the reference view for training stability and faster convergence.

\section{Preliminary of State Space Models}
Utilizing ideas from the control theory~\cite{glasser1985control}, the integration of linear state space equations with deep learning has been widely employed to tackle the modeling of sequential data. The promising property of linearly scaling with
sequence length in long-range dependency modeling has attracted great interest from searchers. Pioneered by LSSL~\cite{gu2021combining} and S4~\cite{gu2021efficiently}, which utilize linear state space equations for sequence data modeling, follow-up works mainly focus on memory efficiency~\cite{gu2021efficiently}, fast training speed~\cite{gu2022train,gu2022parameterization} and better performance~\cite{mehta2022long,wang2023selective}. More recently, Mamba~\cite{mamba} integrates a selective mechanism and efficient hardware design,
outperforms Transformers~\cite{vaswani2017attention} on natural language and enjoys linear scaling with
input length. Building on the success of Mamba, Vision Mamba~\cite{visionmamba} and VMamba~\cite{vmamba} leverage the bidirectional Vim Block and the Cross-Scan Module respectively to gain data-dependent global visual context for visual representation; U-Mamba~\cite{ma2024u} and Vm-unet~\cite{ruan2024vm} further bring Mamba into the field of medical image segmentation. PointMamba~\cite{liang2024pointmamba} and Point Cloud Mamba ~\cite{zhang2024point} adapt Mamba for point cloud understanding through reordering and serialization strategy.

\noindent\textbf{State Space Models (SSMs)}~\cite{gu2021efficiently} have emerged as a powerful tool for modeling and analyzing complex physical systems, particularly those that exhibit linear time-invariant (LTI) behavior. The core idea behind SSMs is to represent a system using a set of first-order differential equations that capture the dynamics of the system's state variables. This representation allows for a concise and intuitive description of the system's behavior, making SSMs well-suited for a wide range of applications.
The general form of an SSM can be expressed as follows:
\begin{equation}
\begin{aligned}
\dot{h}(t) &= Ah(t) + Bx(t), \\
y(t) &= Ch(t) + Dx(t).
\end{aligned}
\end{equation}
where $h(t)$ denotes the state vector of the system at time $t$, while $\dot{h}(t)$ denotes its time derivative. The matrices $A$, $B$, $C$, and $D$ encode the relationships between the state vector, the input signal $x(t)$, and the output signal $y(t)$. These matrices play a crucial role in determining the system's response to various inputs and its overall behavior.

One of the challenges in applying SSMs to real-world problems is that they are designed to operate on continuous-time signals, whereas many practical applications involve discrete-time data. To bridge this gap, it is necessary to discretize the SSM, converting it from a continuous-time representation to a discrete-time one.
The discretized form of an SSM can be written as:
\begin{equation}
\begin{aligned}
h_k &= \bar{A}h_{k-1} + \bar{B}x_k, \\
y_k &= \bar{C}h_k + \bar{D}x_k.
\end{aligned}
\end{equation}
Here, $k$ represents the discrete time step, and the matrices $\bar{A}$, $\bar{B}$, $\bar{C}$, and $\bar{D}$ are the discretized counterparts of their continuous-time equivalents. The discretization process involves sampling the continuous-time input signal $x(t)$ at regular intervals, with a sampling period of $\Delta$. This leads to the following relationships between the continuous-time and discrete-time matrices:
\begin{equation}
\begin{aligned}    
\bar{A} &= (I - \Delta/2 \cdot A)^{-1}(I + \Delta/2 \cdot A), \\
\bar{B} &= (I - \Delta/2 \cdot A)^{-1}\Delta B, \\
\bar{C} &= C.
\end{aligned}
\end{equation}

\noindent\textbf{Selective State Space Models}~\cite{gu2023mamba} are proposed to address the limitations of traditional SSMs in adapting to varying input sequences and capturing complex, input-dependent dynamics. The key innovation in Selective SSMs is the introduction of a selection mechanism that allows the model to efficiently select data in an input-dependent manner, enabling it to focus on relevant information and ignore irrelevant inputs. The selection mechanism is implemented by parameterizing the SSM matrices $\bar{B}$, $\bar{C}$, and $\Delta$ based on the input $x_k$. This allows the model to dynamically adjust its behavior depending on the input sequence, effectively filtering out irrelevant information and remembering relevant information indefinitely.

\section{Limitations}
While Gamba achieves remarkable speed and promising results in 3D reconstruction, there are still some limitations. Firstly, the reconstruction quality is highly dependent on the input image; if the input lacks sufficient geometric information, Gamba struggles to produce accurate reconstructions. Secondly, the texture of the reconstructed back view is often smoothed, as it may significantly differ from what is visible in the input image. This issue stems from Gamba being trained under a prediction-only paradigm rather than as a generative model, which limits its ability to infer unseen views of 3D objects. In future, we might explore incorporating a diffusion training to enhance Gamba's generative capabilities. Thirdly, the scalability of the Gamba model remains underexplored. Currently, the model utilizes only 14 blocks, with its parameters amounting to merely about $10\%$ of those employed in existing large reconstruction models. We envision this work as paving the way for alternative approaches in amortized 3D reconstruction, inviting further exploration into scalable architectures.

% \begin{figure}[h!]
%     \centering
%     \includegraphics[width=0.8\linewidth]{figures/gamba_appendix.pdf}
%     \caption{More qualitative results.}
%     \label{fig:app_result}
% \end{figure}

% \section{Licenses}
% Datasets:
% \begin{itemize}
%     \item Objaverse~\cite{deitke2023objaverse}: ODC-By v1.0 license
% \end{itemize}

% Pre-trained models:
% \begin{itemize}
%     \item DINOv2~\cite{dinov2}: Apache License 2.0
%     \item SAM~\cite{kirillov2023segany}: Apache-2.0 license
% \end{itemize}

% \bibliographystyle{ieeenat_fullname}
% \bibliographystyle{unsrt}
% \bibliography{gamba}

\clearpage

{
\small
\bibliographystyle{plain}
\bibliography{gamba}
}

\end{document}